\documentclass{article}

\usepackage{graphicx}
\usepackage{subfigure}
\usepackage{amssymb}
\usepackage{amsthm}
\usepackage{amsmath}
\usepackage{yfonts}
\usepackage{epsf}
\usepackage{xspace}
\usepackage{xcolor}
\usepackage{algorithm}
\usepackage{algpseudocode}
\usepackage{comment}
\usepackage{csquotes}
\usepackage{bookmark}
\usepackage[inline]{enumitem}
\usepackage{etoolbox}
\usepackage{float}
\usepackage{authblk}

\usepackage[numbers]{natbib}

\theoremstyle{plain}
\newtheorem{theorem}{Theorem}[section]

\theoremstyle{definition}
\newtheorem{definition}[theorem]{Definition}
\newtheorem*{definition*}{Definition} 

\theoremstyle{remark}

\numberwithin{theorem}{section}

\title{Reinforcement learning with timed constraints for robotics motion planning}

\author[1]{Zhaoan Wang}
\author[2]{Junchao Li}
\author[3]{Mahdi Mohammad}
\author[1]{Shaoping Xiao}

\affil[1]{Department of Mechanical Engineering, Iowa Technology Institute, University of Iowa, Iowa City, IA, 52242, USA}
\affil[2]{Talus Renewables, Inc. Austin, TX, 78754, USA}
\affil[3]{Department of Industrial, Electronic, and Mechanical Engineering, Roma Tre University, Rome, 00146, Italy}

\date{\today}

\begin{document}

\maketitle

\begin{abstract}
Robotic systems operating in dynamic and uncertain environments increasingly require planners that satisfy complex task sequences while adhering to strict temporal constraints. Metric Interval Temporal Logic (MITL) offers a formal and expressive framework for specifying such time-bounded requirements; however, integrating MITL with reinforcement learning (RL) remains challenging due to stochastic dynamics and partial observability. This paper presents a unified automata-based RL framework for synthesizing policies in both Markov Decision Processes (MDPs) and Partially Observable Markov Decision Processes (POMDPs) under MITL specifications. MITL formulas are translated into Timed Limit-Deterministic Generalized Büchi Automata (Timed-LDGBA) and synchronized with the underlying decision process to construct product timed models suitable for Q-learning. A simple yet expressive reward structure enforces temporal correctness while allowing additional performance objectives. The approach is validated in three simulation studies: a $5 \times 5$ grid-world formulated as an MDP, a $10 \times 10$ grid-world formulated as a POMDP, and an office-like service-robot scenario. Results demonstrate that the proposed framework consistently learns policies that satisfy strict time-bounded requirements under stochastic transitions, scales to larger state spaces, and remains effective in partially observable environments, highlighting its potential for reliable robotic planning in time-critical and uncertain settings.    
\end{abstract}

\section{Introduction}

With rapid advances in robotics, sensing technologies, and computational methods, modern robotic systems are increasingly expected to operate autonomously in complex and dynamic environments, from human-centric indoor spaces to large-scale industrial or outdoor settings \cite{reed2021, thomas2022}. As robots undertake more sophisticated responsibilities, including assisting humans, navigating transportation networks, or performing tasks in uncertain and unstructured environments, the demand for reliable, robust, and scalable motion-planning frameworks has grown substantially. These rising expectations highlight the need for planning algorithms capable of ensuring safe, efficient, and goal-directed behavior under diverse operational constraints.

Motion planning is fundamentally concerned with determining feasible trajectories that guide a robot from an initial state to a goal while avoiding obstacles and respecting system constraints \cite{reed2021, plaku2016}. Beyond geometric collision avoidance, modern robotic systems must contend with uncertainty arising from sensing noise, actuation errors, and dynamic or partially known environments \cite{morteza2011}. Such challenges appear across a wide range of domains, including robotics \cite{thomas2022}, autonomous driving \cite{david2015}, and human-centered environments such as homes and hospitals \cite{reed2021}. As robotic tasks become more complex, planning methods are increasingly required to satisfy sequential objectives, deadlines, and temporal requirements rather than simple point-to-point navigation.

To address these challenges, temporal logic has emerged as a principled framework for specifying high-level robotic tasks. Linear Temporal Logic (LTL) enables the expression of sequencing, persistence, and conditional behaviors using a compact set of logical and temporal operators \cite{baier2008}. As a result, LTL has been widely adopted for motion planning and control synthesis in robotics \cite{plaku2016, barbosa2019}. Temporal logic specifications allow planners to reason about what tasks should be accomplished and in what order, rather than focusing solely on geometric feasibility.

Despite significant progress, generating safe and feasible trajectories under real-time constraints remains challenging\cite{cai2023safe, li2025reinforcement}. González \textit{et al.} \cite{david2015} surveyed motion-planning algorithms for autonomous-vehicle decision-making and emphasized that, across applications, the core problem remains generating safe, feasible trajectories from start to goal. A key difficulty is the requirement for timely decision-making in highly dynamic environments, where limited computational time can hinder the synthesis of safe and reactive motion plans. Real-time motion planning, therefore, remains an open and active research area.

To better capture continuous-time behaviors and robustness, Signal Temporal Logic (STL) has been proposed as an extension of temporal logic to continuous signals \cite{maler2004}. STL allows the specification of complex spatial and temporal requirements and has been applied to multi-agent motion planning by optimizing temporal robustness margins \cite{joris2023}. Related work by Sun \textit{et al.} \cite{sun2022} demonstrated the use of STL for cooperative multi-agent planning, enabling autonomous task execution while satisfying temporal constraints. These studies highlight the expressive power of temporal logic for encoding rich robotic behaviors in continuous domains.

Temporal logic has also been extensively studied within probabilistic planning frameworks. Lahijanian \textit{et al.} \cite{morteza2011} modeled robotic motion using a Markov Decision Process (MDP) and synthesized control policies satisfying probabilistic computation tree logic (PCTL) specifications. LTL, in particular, is valuable for expressing complex missions that extend beyond point-to-point navigation \cite{barbosa2019}. By combining logical and temporal operators, LTL enables specification of rich tasks \cite{plaku2016}.  Cai et al. \cite{mingyu2021} integrated LTL with MDPs featuring unknown transition probabilities, using reinforcement learning and relaxed product MDPs to handle environmental uncertainty. Their later work \cite{mingyu2023} employed Probabilistic Labeled MDPs (PL-MDPs) and Limit-Deterministic Büchi Automata (LDBA) to quantify and mitigate task violations when specifications are infeasible.

To further address uncertainty in sensing and perception, temporal logic has been studied in conjunction with Partially Observable Markov Decision Processes (POMDPs). POMDPs provide a principled framework for planning under stochastic dynamics and incomplete state information. Several works have integrated LTL with POMDPs to specify high-level mission objectives while reasoning over belief space evolution. Li \textit{et al.} \cite{li2023model} proposed a model-based planning framework for POMDPs with LTL specifications, combining belief-space representations with automaton-based encoding of temporal logic tasks to synthesize policies that maximize the probability of satisfying mission requirements under partial observability.

Recent progress in automata-based synthesis has further influenced LTL-based planning. Sickert \textit{et al.} \cite{sickert2016} introduced a direct translation from LTL to Limit-Deterministic Büchi Automata (LDBA), avoiding the costly intermediate nondeterministic automaton. Li \textit{et al.} \cite{junchao2024} leveraged Limit-Deterministic Generalized Büchi Automata (LDGBA) to find optimal policies in POMDPs with complex LTL tasks, while Oura \textit{et al.} \cite{oura2020} augmented LDGBA structures to mitigate reward sparsity and improve optimal control learning. 

While LTL provides expressive power for sequencing and logical dependencies, its linear-time semantics make it difficult to encode explicit real-time constraints without significantly increasing algorithmic complexity \cite{plaku2016}. This limitation motivates the use of temporal logics that directly incorporate quantitative timing information. Metric Temporal Logic (MTL), originally proposed by Koymans \cite{ron1990}, extends LTL by introducing metric temporal operators that allow the specification of time-bounded requirements \cite{fu2015}. A widely used fragment, Metric Interval Temporal Logic (MITL), enables the expression of event ordering with explicit lower and upper temporal bounds and has been applied extensively in real-time and safety-critical robotic systems \cite{linard2024, niu2023}.

Several works have explored MITL-based planning and control synthesis. Ferrere et al. \cite{ferrere2019} translated MITL formulas into timed automata for model checking and runtime monitoring. Nikou \textit{et al.} \cite{nikou2017} proposed decentralized controller synthesis for multi-agent systems under MITL constraints, while Niu \textit{et al.}. \cite{niu2023} enforced MITL specifications in adversarial cyber–physical systems using deterministic timed Büchi automata. Other studies have integrated MITL with stochastic and reinforcement-learning frameworks. Andersson \textit{et al.} \cite{andersson2017} synthesized motion plans for multi-agent systems under continuous-time constraints by constructing local and global Büchi Weighted Transition Systems (BWTS). Fu \textit{et al.} \cite{fu2015} addressed continuous-time stochastic systems with MITL objectives using Markov-chain approximations, while Lin \textit{et al.} \cite{lin2020} proposed a modular Q-learning framework incorporating MITL monitors to enable adaptation under time constraints. Li \textit{et al.} \cite{li2021} further extended MITL with probabilistic semantics, translating specifications into stochastic timed automata suitable for game-theoretic environments.

The integration of MITL with reinforcement learning enables strategies that aim not only to maximize expected rewards but also to adhere rigorously to time-sensitive and safety-critical logic specifications. Such approaches significantly enhance the reliability and effectiveness of robotic systems operating in complex real-world environments.

In this study, we develop a reinforcement learning framework for synthesizing policies in both MDPs and POMDPs subject to time-bounded task specifications. Task requirements are expressed using MITL, which enables the explicit specification of lower and upper temporal bounds on event occurrences. Such constraints naturally arise in robotic navigation and service tasks, where actions must be completed within prescribed time windows. To enforce MITL specifications during learning, we translate them into a Timed Limit-Deterministic Generalized Büchi Automaton (Timed-LDGBA) and synchronize the automaton with the system dynamics. This construction explicitly tracks temporal progress and detects violations such as early or late visits through transitions to sink states. A simple reward structure is then defined over the automaton-augmented model, assigning positive rewards only when all temporal constraints are satisfied within their specified intervals, while remaining compatible with standard Q-learning methods. Additional performance objectives, such as movement costs, can be incorporated without altering the temporal semantics.

The main contributions of this work are twofold. First, we propose a unified automata-based reinforcement learning framework that integrates MITL with learning through timed automata, enabling the systematic enforcement of real-time temporal constraints. Unlike existing MITL-based reinforcement learning approaches that rely on generic timed automata or runtime monitors primarily for task tracking in fully observable MDPs, our framework constructs a Timed-LDGBA explicitly tailored for learning. This automaton preserves MITL semantics while reducing nondeterminism and supporting multiple accepting conditions, making it well-suited for reward shaping and scalable policy synthesis.

Second, the proposed framework seamlessly extends to partially observable environments. By synchronizing the Timed-LDGBA with both MDPs and POMDPs, we obtain a unified product timed model that supports reinforcement learning under full and partial observability within a single modeling paradigm. While prior work has explored reinforcement learning under MITL specifications by augmenting MDPs with timed automata or monitors, these approaches are restricted to fully observable settings and do not address learning under partial observability \cite{lin2020, xu2019transfer}. To the best of our knowledge, this work is the first to provide an automata-based reinforcement learning framework that integrates MITL specifications with POMDPs.

This paper is organized as follows. Section 2 introduces the formal foundations of reinforcement learning (RL), including Deep Q-learning (DQN), MDPs, and POMDPs. Section 3 presents MITL, Timed-LDGBA, and the construction of the corresponding product timed MDPs and POMDPs. Section 4 reports simulation results in grid-world environments, including a 5×5 workspace, a 10×10 workspace, and an office-like environment. Finally, Section 5 concludes the paper and discusses directions for future research.

\section{Reinforcement Learning}

RL is a framework for sequential decision making in which an agent interacts with an environment over time to maximize cumulative reward. At each time step, the agent observes a state (or, more generally, an observation), selects an action according to a policy, and then receives a scalar reward and a new state from the environment. The objective is to learn a policy that maximizes the expected discounted return, typically by modeling the environment as an MDP. Unlike supervised learning, which relies on labeled input--output pairs, RL must cope with delayed and stochastic feedback as well as the exploration and exploitation trade-off.

In many real-world applications, RL has been used to intelligently synthesize control policies from data through interaction with environments that are often accurately modeled as POMDPs. For example, in control and robotics, RL is used to learn policies for locomotion, manipulation, and autonomous navigation\cite{li2023model}. In agricultural management, RL can be employed to generate fertilization \cite{wang2024learning} and irrigation \cite{wang2025reinforcement} policies that adapt to soil conditions, weather forecasts, and crop growth stages in order to optimize yield and resource usage. Similarly, in water resources management, RL can be used to control dam operations during flooding events, learning release strategies that balance flood mitigation, reservoir safety, and downstream risk \cite{tofighi2025data}.

\subsection{Markov Decision Process}

An MDP provides the standard mathematical framework for modeling sequential decision-making problems under full observability.

\begin{definition}[MDP]\label{def:MDP}
An MDP is defined as a tuple  
$\mathcal{M} = (S, A, T, s_0, R, \Pi, L)$, where:
\begin{itemize}
\item $S = \{s_1, \ldots, s_n\}$ is a finite set of states.
\item $A = \{a_1, \ldots, a_m\}$ is a finite set of actions, where $A(s)$ denotes the actions available at state $s$.
\item $T : S \times A \times S \rightarrow [0,1]$ is the transition probability function, satisfying $\sum_{s'} T(s,a,s') = 1$.
\item $s_0 \in S$ is the initial state.
\item $R : S \times A \times S \rightarrow \mathbb{R}$ is the reward function.
\item $\Pi$ is a set of atomic propositions.
\item $L : S \rightarrow 2^{\Pi}$ is a labeling function.
\end{itemize}
\end{definition}

In an MDP, the agent fully observes the current state $s$, selects an action $a \in A(s)$, transitions to a successor state $s'$ according to $T(s,a,s')$, and receives a scalar reward $R(s,a,s')$. The objective is to learn a policy that maximizes the expected cumulative discounted reward. $L(s)$ describes the event (represented by atomic propositions $\Pi$) occurring at state $s$ for automata-based RL\cite{junchao2024}. 

Building on this framework, deep Q-learning extends classical Q-learning~\cite{watkins1992q} to environments with large or continuous state spaces. Instead of maintaining a tabular representation of the state--action value function, the DQN~\cite{mnih2013playing} employs a deep neural network (DNN) parameterized by weights $\theta$ to approximate the Q-function. When applied to an MDP, the DNN takes the fully observable state as input and outputs Q-values for all available actions. As a representative method in deep reinforcement learning (DRL), DQN enables efficient learning in settings where tabular methods are infeasible.

A DQN architecture typically consists of two networks: an \emph{evaluation network} $Q_e(s,a;\theta_e)$, which is updated at every learning step, and a \emph{target network} $Q_t(s,a;\theta_t)$, whose parameters are held fixed for a period and periodically updated by copying the evaluation network parameters (i.e., $\theta_t \leftarrow \theta_e$). To improve learning stability, DQN employs \emph{experience replay}~\cite{lin1992self}, where experience tuples $(s,a,s',R)$ are stored in a replay buffer and randomly sampled to break temporal correlations. The Bellman update used in DQN is given by
\begin{equation} \label{eq:DQN}
Q_{\text{new}}(s,a) = Q_e(s,a;\theta_e) + \alpha \Big[ R(s,a,s') + \gamma \max_{a' \in A} Q_t(s',a';\theta_t) - Q_e(s,a;\theta_e) \Big]
\end{equation}
where $\alpha$ is the learning rate, enhancing the efficiency and stability of Q-value convergence. The discount factor, $\gamma \in [0,1]$, weights the contribution of future rewards relative to immediate rewards in the agent’s decision-making process. Under the assumption of full observability, DQN effectively leverages the MDP structure by directly mapping states to action values, enabling the learning of near-optimal policies in high-dimensional environments.

\subsection{Partially Observable Markov Decision Process}

In many environments, the agent does not have full access to the underlying state. Instead, it receives only partial and noisy observations, making the environment naturally modeled by the POMDP framework.

\begin{definition}[POMDP]\label{def:POMDP}
A POMDP is a tuple $\mathcal{M} = (S, A, T, s_0, R, O, \Omega, \Pi, L)$, where $S$, $A$, $T$, $s_0$, $R$, $\Pi$, and $L$ are defined as in the MDP~(Definition \ref{def:MDP}), and:
\begin{itemize}
    \item $O = \{o_1, \ldots, o_z\}$ is a finite observation set.
    \item $\Omega : S \times A \times O \to [0,1]$ is the observation probability function satisfying $\sum_{o \in O} \Omega(s', a, o) = 1$ for all $s' \in S$ and $a \in A$.
\end{itemize}
\end{definition}

Because the true state $s$ is hidden, the agent must rely on observations and their history. After executing action $a$ and transitioning to a hidden state $s'$, the agent receives observation $o$ with probability $\Omega(s',a,o)$. The goal remains to maximize the expected discounted return:
\begin{equation}\label{chap2:eq:expReturn}
U(s) = \mathbb{E} \left[\sum_{t=0}^{\infty} \gamma^t R(s_t, a_t, s_{t+1}) \,\bigm|\, s_{0} = s\right].
\end{equation}

In the fully observable MDP setting, a DQN approximates the optimal action-value function $Q^*(s,a)$ with a neural network $Q(s,a;\theta)$ that directly takes the Markov state $s_t$ as input. Given a transition $(s_t, a_t, r_t, s_{t+1})$, the standard DQN target is
\begin{equation}
y_t = r_t + \gamma \max_{a' \in A} Q(s_{t+1}, a'; \theta^-),
\end{equation}
which is valid because the current state $s_t$ is a sufficient statistic of the past (the Markov property).

In a POMDP, however, the agent observes only $o_t$ instead of the true state $s_t$, so a single observation is generally non-Markovian. A naive DQN that learns $Q(o_t, a_t; \theta)$ can therefore be unstable or significantly suboptimal. A common DQN variant for POMDPs augments the Q-network with a recurrent neural network (e.g., an LSTM or GRU) that maintains a hidden state:

\begin{equation}
h_t = f_{\theta_h}(h_{t-1}, o_t),
\end{equation}
and parameterizes the action-value function as
\begin{equation}
Q(o_{0:t}, a_t; \theta) \approx Q(h_t, a_t; \theta).
\end{equation}
The temporal-difference (TD) target keeps the same functional form as in the MDP case, but is now computed using the recurrent state $h_{t+1}$. This hidden state summarizes the observation history, allowing the network to better handle partial observability. 

\section{Timed Constraints }

\subsection{Metric Interval Temporal Logic}
Metric Interval Temporal Logic (MITL)~\cite{alur1996} is interpreted over discrete time in this work.
\begin{definition}[MITL]\label{def:MITL}
Let $\mathbb{N}_0 = \{0,1,2,\dots\}$ denote the set of non-negative integers, and let $\Pi$ be a finite set of atomic propositions, and let $2^\Pi$ denote its power set. 

The set of MITL formulas is defined by the following grammar~\cite{niu2023}:

\begin{equation}\label{eq:MITL}
  \phi ::= \text{True} \mid \pi \mid \phi_1 \land \phi_2
        \mid \lnot \phi \mid \bigcirc \phi \mid \phi_1 \mathcal{U}_I \phi_2 ,
\end{equation}

where $\pi \in \Pi$ and $I \subseteq \mathbb{N}_0$ a discrete time interval. A typical interval is of the form $I = [\tau_1,\tau_2] \cap \mathbb{N}_0$ with $\tau_1, \tau_2 \in \mathbb{N}_0$ and $\tau_1 \le \tau_2$.

An infinite word over $2^\Pi$ is a sequence 
$w = w_0 w_1 w_2 \dots$ with $w_i \in 2^\Pi$ for all $i \in \mathbb{N}_0$.
Formulas are interpreted at positions $i \in \mathbb{N}_0$; we write
$(w,i) \models \phi$ to mean that $\phi$ holds at position $i$ of $w$.
The satisfaction relation is defined inductively as follows:
\begin{equation}
\arraycolsep=1.4pt
\begin{array}{lcl}
(w,i) \models \text{True} \\[2pt]
(w,i) \models \pi &\Leftrightarrow& \pi \in w_i \\[2pt]
(w,i) \models \phi_1 \land \phi_2 &\Leftrightarrow&
    (w,i) \models \phi_1 \text{ and } (w,i) \models \phi_2 \\[2pt]
(w,i) \models \lnot \phi &\Leftrightarrow&
    (w,i) \not\models \phi \\[2pt]
(w,i) \models \bigcirc \phi &\Leftrightarrow&
    (w,i+1) \models \phi \\[2pt]
(w,i) \models \phi_1 \mathcal{U}_I \phi_2 &\Leftrightarrow&
    \exists j \ge i \text{ such that } (j-i) \in I,\ (w,j) \models \phi_2,\\
&& \text{and } \forall k \text{ with } i \le k < j,\ (w,k) \models \phi_1 .
\end{array}
\end{equation}

The derived temporal operators \emph{constrained eventually} and
\emph{constrained always} are defined by:
\begin{equation}
\arraycolsep=1.4pt
\begin{array}{lcl}
\diamondsuit_I \phi &\equiv& \text{True } \mathcal{U}_I \phi, \\[2pt]
\square_I \phi &\equiv& \lnot (\diamondsuit_I \lnot \phi).
\end{array}
\end{equation}
Intuitively, $\diamondsuit_I \phi$ means that $\phi$ becomes true at some
future position whose distance from the current position lies in $I$, and
$\square_I \phi$ means that $\phi$ holds at all such positions.
\end{definition}

\subsection{Timed Limit-Deterministic Generalized B\"uchi Automata (T-LDGBA)}

\begin{definition}[T-LDGBA]\label{def:T-LDGBA}
A T-LDGBA is an extension of LDGBA\cite{cai2023} by incorporating timed constraints into the acceptance conditions, and it is denoted as a tuple \(\mathcal{A} = (Q, \Sigma, \mathcal{I}, \varphi, \delta, q_0, \mathcal{F})\), where the finite set of time intervals $\mathcal{I}$ is derived directly from the temporal bounds specified in the MITL formula. These intervals provide an automaton-level realization of MITL temporal semantics by constraining the enabling of transitions according to the corresponding time bounds.

\begin{itemize}
    \item \(Q = \{q_1, q_2, \ldots, q_n\}\) is a finite set of states.
    \item \(\Sigma = 2^{\Pi}\) is a finite alphabet or a finite set of input symbols,
    where \(\Pi\) is a set of atomic propositions.
    \item $\mathcal{I} = \{I_1, I_2, ...\}$ is a finite set of time intervals.
    \item $\varphi : Q \times Q \times \mathcal{I} \to $ \{TRUE, FALSE\} is the clock constraint function, which outputs a Boolean value to determine whether an automaton state transition is enabled within a given time interval.
    \item $\delta : Q \times ({\Sigma} \cup {\varepsilon}) \times \mathcal{I} \to 2^{Q}$
          is the transition function.  
          The \(\varepsilon\)-transition is not allowed in the deterministic set
          and is only defined for state transitions from \(Q_N\) to \(Q_D\),
          which do not consume the input alphabet.
    \item \(q_0 \in Q\) is an initial state.
    \item \(\mathcal{F} = \{\mathcal{F}_1, \mathcal{F}_2, \ldots, \mathcal{F}_f\}\)
          is a set of accepting sets with \(\mathcal{F}_i \subseteq Q\),
          \(\forall i \in \{1, \ldots, f\}\).
\end{itemize}

\begin{figure}
\centering
\includegraphics[width=70mm]{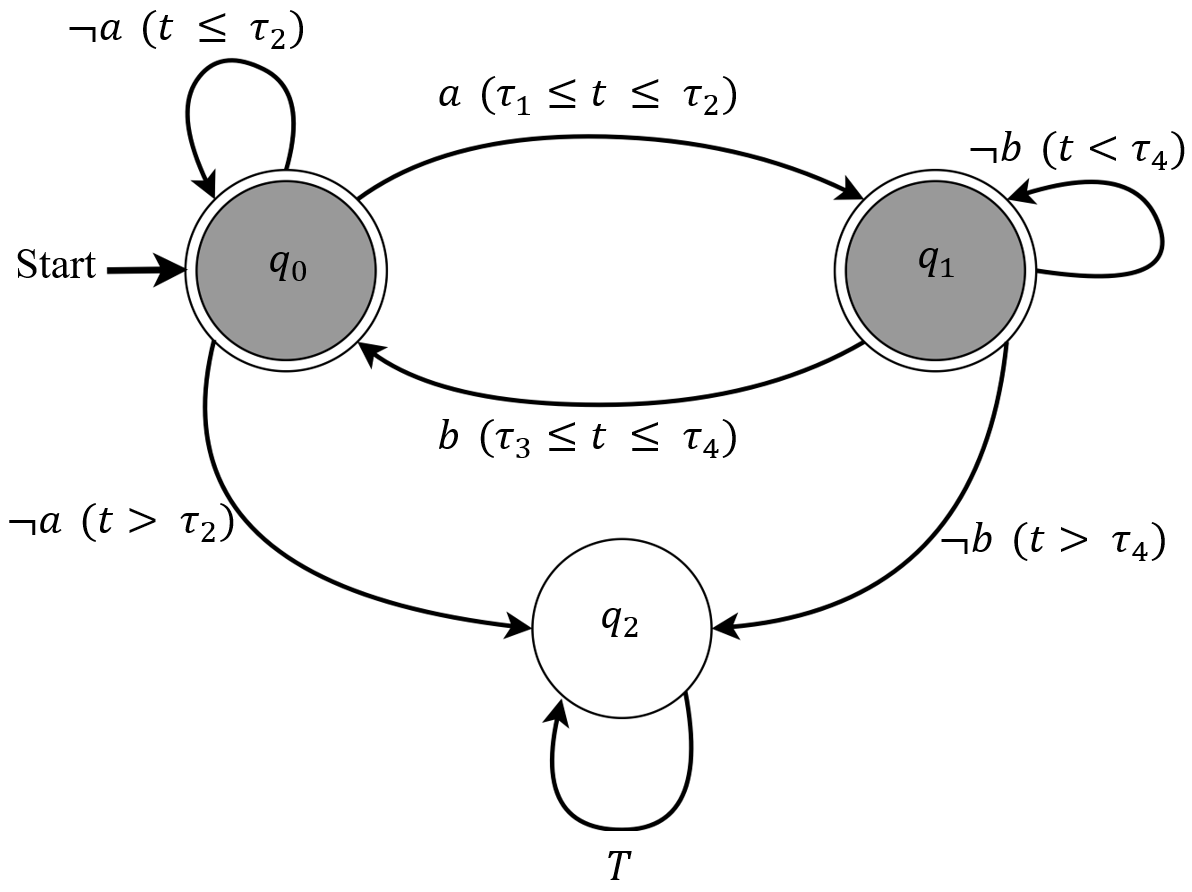}
\centering\caption{An example of a three-state T-LDGBA.}
\label{fig:T-LDGBA_Graph}
\end{figure}

Fig.~\ref{fig:T-LDGBA_Graph} illustrates an example of T-LDGBA $\mathcal{A}$ that accepts the runs that satisfy formula $\phi_{1}= \square\diamondsuit_{[\tau_{\text{1}}, \tau_{\text{2}}]}a \land \square\diamondsuit_{[\tau_{\text{3}}, \tau_{\text{4}}]}b$. The automaton consists of three states $q_0$, $q_1$, and $q_2$, where $q_0$ is the initial state. States $q_0$ and $q_1$ are accepting states, representing the progress toward satisfying the temporal obligations associated with $a$ and $b$, respectively, while $q_2$ is a sink state.

From the initial state $q_0$, where it waits for proposition $a$ to occur within the time interval $[\tau_{1}, \tau_{2}]$. Upon observing $a$ in this interval, it transitions to state $q_1$. If $a$ does not occur before $\tau_{2}$, the automaton moves to the sink state $q_2$. In state $q_1$, the automaton enforces the occurrence of proposition $b$ within $[\tau_{3}, \tau_{4}]$. Observing $b$ within this interval returns the automaton to $q_0$, completing one satisfaction cycle. Failure to observe $b$ before $\tau_{4}$ also leads to $q_2$. 

The sink state $q_2$ is absorbing and represents a permanent violation of the MITL specification. The self-loop on $q_2$ labeled $T$ denotes a tautology, meaning that the transition is unconditionally enabled regardless of the valuation of atomic propositions. Consequently, once the automaton enters $q_2$, it remains there indefinitely, and any run that reaches $q_2$ is rejected.

\end{definition}

\subsection{Product timed MDP/POMDP}

\begin{definition}[Product timed MDP]\label{def:PMDP}
Consequently, a product timed MDP can be generated from an MDP and a T-LDGBA,
defined as a tuple $\mathcal{M}^{\times} = (S^{\times}, A^{\times}, T^{\times}, \mathcal{I}, \varphi, s_0^{\times}, R^{\times}, \mathcal{F}^{\times})$, where:

\begin{itemize}
    \item \(S^{\times} = S \times Q\) is the finite set of labeled states, i.e.,
    \(s^{\times} = \langle s, q \rangle \in S^{\times}\) where \(s \in S\) and \(q \in Q\).

    \item \(A^{\times} = A \cup \{\varepsilon\}\) is the set of actions.

    \item \(T^{\times} : S^{\times} \times A^{\times} \times \mathcal{I} \times S^{\times} \rightarrow [0,1]\)
    is the constrained transition function, specifically,
    \[
      T^{\times}(s^{\times}, a^{\times}, s^{\times'}, I) =
      \begin{cases}
         T(s, a, s') & 
         \text{if } a \in A,\; l \in L(s'),\; q' = \delta(q, l),\; \varphi(q,q',I) = \text{TRUE}, \\[4pt]
         1 & 
         \text{if } a^{\times} \in \{\varepsilon\},\; q' \in \delta(q, \varepsilon),\; s' = s, \\[4pt]
         0 & \text{otherwise}.
      \end{cases}
    \]

    \item \(s_0^{\times} = \langle s_0, q_0 \rangle \in S^{\times}\) is the initial state, where
    \(s_0 \in S\) and \(q_0 \in Q\).

    \item \(R^{\times} : S^{\times} \rightarrow \mathbb{R}\) is the reward function, and
    \[
      R^{\times}(s^{\times}) =
      \begin{cases}
         R(s) & \text{if } l \in L(s),\; q' = \delta(q, l) \in \mathcal{F}_i,\; \mathcal{F}_i \in \mathcal{F},\\[4pt]
         0 & \text{otherwise}.
      \end{cases}
    \]

    \item \(\mathcal{F}^{\times} = \{ \mathcal{F}^{\times}_1, \ldots, \mathcal{F}^{\times}_f \}\) is the set of accepting sets,
    where \(\mathcal{F}^{\times}_i = \{ \langle s, q \rangle \mid s \in S,\; q \in \mathcal{F}_i \}\),
    for \(i = 1, \ldots, f\).
\end{itemize}
\end{definition}

\begin{definition}[Product timed POMDP]\label{def:PPOMDP}
The product timed POMDP of a POMDP and a T-LDGBA can be written as  
\[
\mathcal{P}^{\times} =
(S^{\times}, A^{\times}, T^{\times}, \mathcal{I}, \varphi, s_0^{\times},
R^{\times}, O, \Omega^{\times}, \mathcal{F}^{\times}),
\]
where \(S^{\times}, A^{\times}, T^{\times}, s_0^{\times},
R^{\times},\) and \(\mathcal{F}^{\times}\) are the same as defined in the product timed MDP.  
Additionally:

\begin{itemize}
    \item \(\Omega^{\times} : S^{\times} \times A^{\times} \times \mathcal{I} \times O \to [0,1]\)
    is the constrained observation function, specifically,
    \[
      \Omega^{\times}(s^{\times}, a^{\times}, o, I) =
      \begin{cases}
         \Omega(s', a, o) &
            \text{if } l \in L(s'),\;
            q' = \delta(q, l),\;
            a \in A,\;
            \varphi(q, q', I) = \text{TRUE}, \\[4pt]
         0 & \text{otherwise}.
      \end{cases}
    \]
\end{itemize}
\end{definition}

The product POMDP constructed with the Timed-LDGBA is itself a valid POMDP, and an optimal policy is defined as one that maximizes the expected return. Given a state trajectory $s_0,s_1,\dots$ in the underlying POMDP, the associated sequence of state labels and elapsed time induces a corresponding run of the Timed-LDGBA through its transition function and clock constraints. By augmenting the POMDP state with the automaton state and clock valuation, the product POMDP can be viewed as an equivalent POMDP with an expanded state space that jointly captures physical dynamics, logical progression, and temporal feasibility.

Previous studies have addressed MDPs with LTL specifications by constructing a product MDP and applying model-checking techniques to synthesize correct policies~\cite{cai2021modular}. A similar strategy can be utilized for the product POMDP in this study: the optimal policy for the original POMDP can therefore be induced from an optimal policy on the product POMDP. Since the MITL specifications are explicitly encoded in the Timed-LDGBA component of the product, all executions generated by the induced policy satisfy the specified time-bounded temporal requirements by construction.

In practice, explicitly constructing the full product timed MDP/POMDP may be computationally prohibitive due to the enlarged state space. Consequently, in our simulations we adopt an on-the-fly product construction, where automaton transitions and constraint checks are performed online during learning.

\section{Simulations}

\subsection{Case 1: 5$\times$5 grid world workspace as an MDP}

\begin{figure}
\centering
\includegraphics[width=50mm]{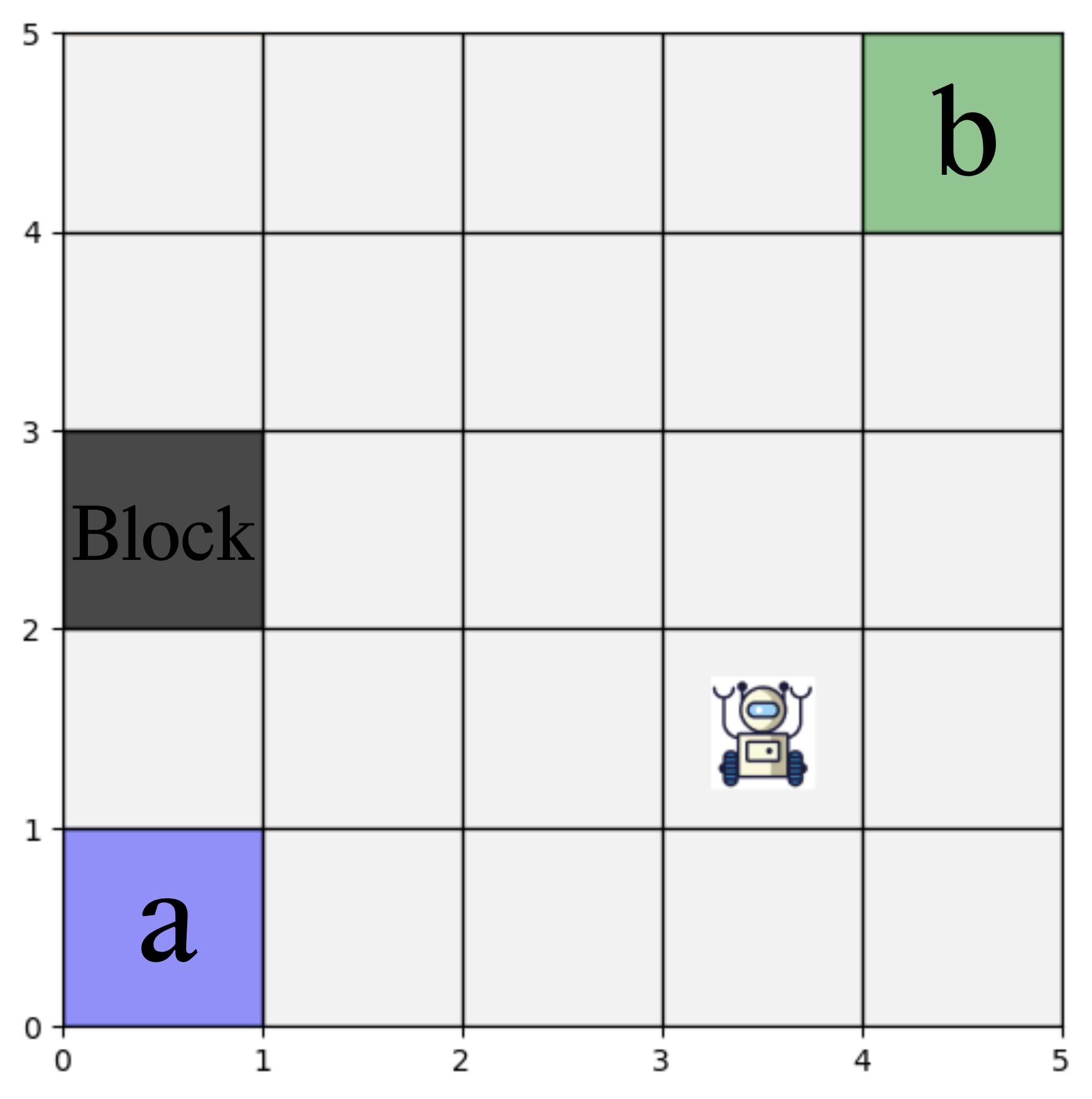}
\centering\caption{A 5 $\times$ 5 grid world, with two labels (\textbf{a}) in blue and (\textbf{b}) in green and one \textbf{block} in black.}
\label{fig:grid_world_5_by_5}
\end{figure}

In the first case, we consider a 5 $\times$ 5 grid world workspace, shown in Figure~\ref{fig:grid_world_5_by_5}. Within this environment, one state is labeled ‘$a$’ (blue) and another ‘$b$’ (green), representing two distinct events. The agent, which can be interpreted as a mobile robot operating within the grid, is required to visit the ‘$a$’ state and subsequently the ‘$b$’ state in a recursive manner. This environment is modeled as a fully observable MDP, in which the agent has access to its exact grid position, and uncertainty arises solely from stochastic transition dynamics.

At each state, the agent can choose one of five possible actions: $up$, $left$, $down$, $right$, and $stay$. In the study, stochastic transitions are introduced to account for motion uncertainty: the agent moves in the intended direction with probability 0.8, while the remaining probability 0.2 is equally divided between the two possible sideways directions. When the stay action is chosen, the agent remains in its current state with probability 1.0. The agent is not allowed to leave the grid-world boundaries at any time.

In addition, the agent must visit the states labeled ‘$a$’ and ‘$b$’ within specified time intervals, as defined by the MITL specifications. The MITL formula of the specified task is defined as: 
\begin{equation}
    \square\diamondsuit_{I_{1}}a \land \square\diamondsuit_{I_{2}}b = \square\diamondsuit_{[\tau_{\text{1}}, \tau_{\text{2}}]}a \land \square\diamondsuit_{[\tau_{\text{3}}, \tau_{\text{4}}]}b
\end{equation}

Specifically, $I_{1}$ and $I_{2}$ denote the time intervals associated with the agent’s visits to the states labeled ‘$a$’ and ‘$b$’, respectively, where $I_{1} = [\tau_{\text{1}}, \tau_{\text{2}}]$ and $I_{2} = [\tau_{\text{3}}, \tau_{\text{4}}]$. When the MITL formula is translated into a timed automaton, these intervals define the corresponding temporal constraints, as formalized in Section~\ref{def:T-LDGBA}. Accordingly, the agent is required to visit the state labeled ‘$a$’ within the time interval $[\tau_{\text{1}}, \tau_{\text{2}}]$ and the state labeled ‘$b$’ within the interval $[\tau_{\text{3}}, \tau_{\text{4}}]$.

In this case, the temporal bounds are explicitly defined as $\tau_{\text{1}} = 5$, $\tau_{\text{2}} = 10$, $\tau_{\text{3}} = 15$, and $\tau_{\text{4}} = 20$. Accordingly, the agent must visit the state labeled ‘$a$’ between steps 5 and 10, and subsequently the state labeled ‘$b$’ between steps 15 and 20. These intervals establish the temporal windows within which the agent is required to satisfy the task specifications.

The reward structure is designed to strictly enforce the MITL specification through the automaton-augmented model. In particular, rewards are associated with \emph{accepting states} of the product timed MDP, as defined in Definition~\ref{def:PMDP}. When the agent's action causes the T-LDGBA to transition into an accepting state, the corresponding product state $\langle s, q \rangle \in \mathcal{F}^{\times}$ yields a reward of $100$. In this case, the automaton transitions from the initial state $q_0$ to $q_1$ upon the first successful satisfaction of the time-bounded requirement.

Any violation of the temporal constraints—including early or late visits, missed deadlines, or invalid revisits—causes the automaton to transition to a non-accepting sink state, resulting in a reward of $0$. Once the sink state is reached, the automaton remains there indefinitely, and no further rewards can be obtained.

Two experimental scenarios are considered. In the first scenario, the movement cost is disregarded (i.e., the movement penalty is set to zero), granting the agent full freedom to focus on satisfying the time-constrained label visits (‘$a$’ and ‘$b$’). In the second scenario, a movement cost of -5 is introduced to penalize unnecessary movement, encouraging the agent to optimize its trajectory while still adhering to the temporal constraints. This cost-aware formulation promotes a balance between exploration and efficiency, reducing superfluous motion and improving path optimization, particularly when time is not a stringent limiting factor. It is important to note that the incorporation of movement cost does not interfere with the temporal constraints defined within the MITL and corresponding omega automaton framework.

\begin{figure}[H]
\centering
\subfigure[]{\resizebox*{80mm}{!}{\includegraphics{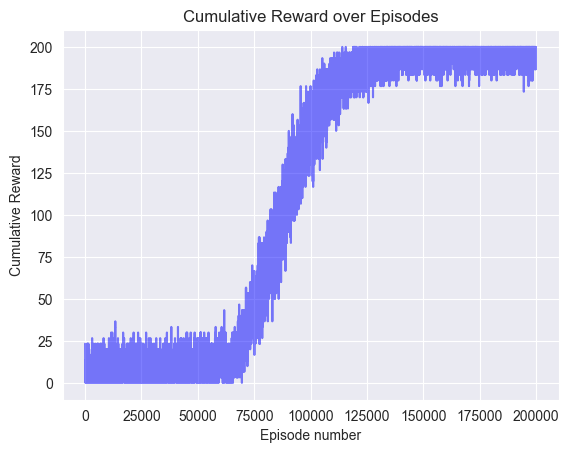}}}
\caption{ The average cumulative reward in the 5 $\times$ 5 grid world without a movement penalty }
\label{fig:5x5m0reward}
\end{figure}

\begin{figure}[H]
\centering
\subfigure[]{\resizebox*{60mm}{!}{\includegraphics{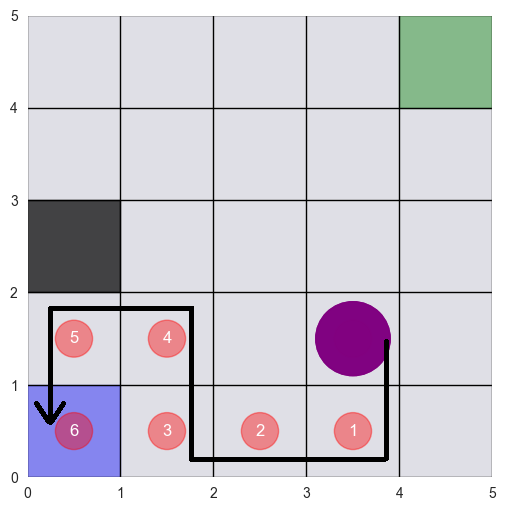}}}
\subfigure[]{\resizebox*{60mm}{!}{\includegraphics{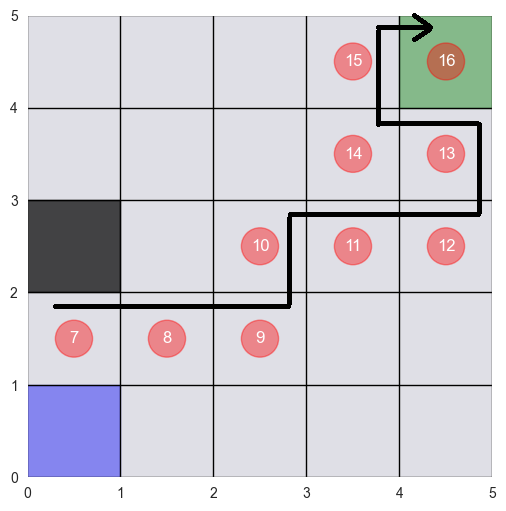}}}
\caption{ A 5 $\times$ 5 grid-world environment illustrating agent trajectories: \textbf{a}) visiting label '$a$', and \textbf{b}) visiting label '$b$'. The Transition is deterministic, and the reward of visiting labels is 100 with no movement penalty. }
\label{fig:5x5m0}
\end{figure}

\begin{figure}[H]
\centering
\subfigure[]{\resizebox*{60mm}{!}{\includegraphics{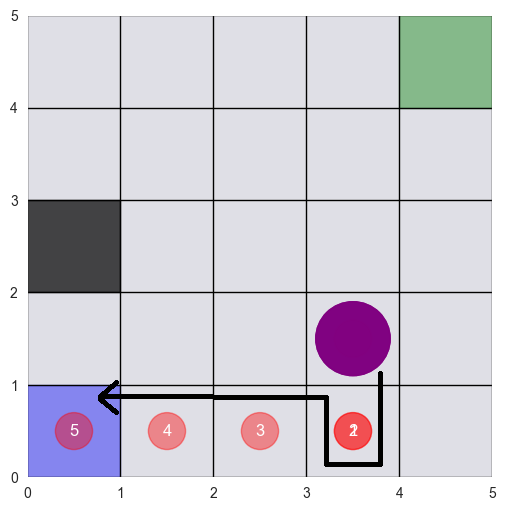}}}
\subfigure[]{\resizebox*{60mm}{!}{\includegraphics{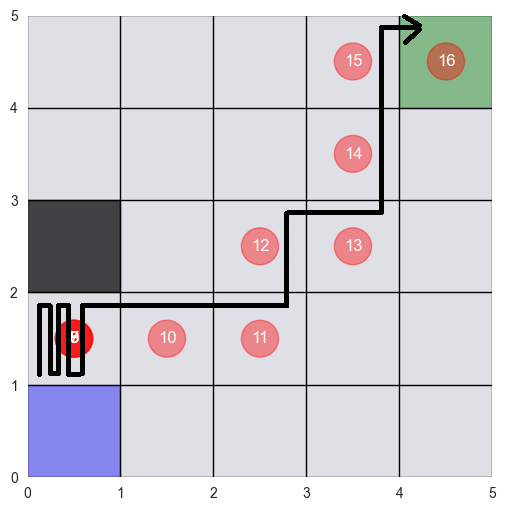}}}
\caption{ A 5 $\times$ 5 grid-world environment illustrating agent trajectories: \textbf{a}) visiting label '$a$', and \textbf{b}) visiting label '$b$'. The Transition is deterministic, and the reward of visiting labels is 100 with the movement penalty. }
\label{fig:5x5m-5}
\end{figure}

Figure~\ref{fig:5x5m0reward} illustrates the cumulative reward, while Figure~\ref{fig:5x5m0} depicts the path-planning results for the first scenario, where the movement cost is ignored, while Figure~\ref{fig:5x5m-5} presents the corresponding results for the second, cost-aware scenario. Both figures demonstrate the agent’s performance in achieving the required label visits (‘$a$’ and ‘$b$’) under their respective conditions.

In this environment, for instance, the agent successfully balances visit timing between two labeled states ($‘a’$ and $‘b’$) and adjusts its navigation patterns depending on whether movement penalties are imposed. When no movement cost is applied, the agent exhibits greater exploratory behavior, selecting paths that prioritize temporal feasibility without regard to path length. In contrast, introducing a negative movement reward encourages the agent to adopt more deliberate and efficient trajectories, reducing unnecessary exploration while still satisfying the prescribed temporal constraints. Notably, the agent continues to achieve the required label visits within their respective time windows in both settings, indicating that the incorporation of auxiliary cost terms does not compromise specification satisfaction. These results demonstrate the flexibility of the proposed framework in preserving the semantics of MITL constraints while simultaneously accommodating additional performance objectives, such as path efficiency, through reward shaping.

\subsection{Case 2: 10$\times$10 grid world workspace as a POMDP}

\begin{figure}
\centering
\includegraphics[width=60mm]{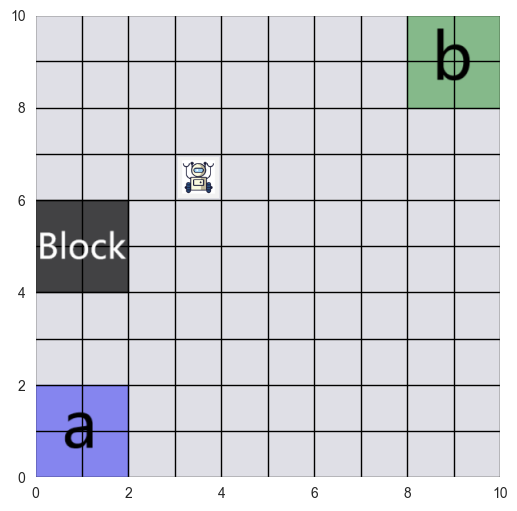}
\centering\caption{A 10 $\times$ 10 grid world, with two labels (\textbf{a}) in blue and (\textbf{b}) in green and one \textbf{block} in black.}
\label{fig:grid_world_10_by_10}
\end{figure}

In the second case study, we consider a 10$\times$10 grid world workspace, as illustrated in Figure~\ref{fig:grid_world_10_by_10}. This environment extends the $5 \times 5$ case to a larger state space and is modeled as a POMDP. As in the previous case, two grid cells are assigned semantic labels: state `$a$' (blue) and state `$b$' (green), representing two distinct events. The agent, which can be interpreted as a mobile robot navigating within the grid, is required to repeatedly visit state `$a$' and subsequently state `$b$' while satisfying time-bounded task specifications.

At each state, the agent can choose one of four possible actions: $up$, $left$, $down$, or $right$. To capture motion uncertainty, stochastic transitions are incorporated: the agent moves in the intended direction with probability 0.8, while the remaining probability (0.2) is equally distributed between the two possible sideways directions. When the $stay$ action is selected, the agent remains in its current state with probability 1.0. Boundary conditions ensure that the agent cannot exit the grid-world at any time.

Unlike the $5 \times 5$ environment, the agent in this case does not have direct access to its true state. Instead, it receives partial observations of the environment, and planning is performed over a belief state, consistent with the POMDP formulation. This setting captures realistic sensing limitations and significantly increases the complexity of the planning problem as the state space grows. Specifically, observations correspond to grid-world states: after transitioning to the next state, the agent observes the true state with probability 0.9, while the remaining probability 0.1 is uniformly distributed among the adjacent states. The agent is provided with full knowledge of the LTL-induced automaton, including its transition function. Consequently, upon reaching a product POMDP state, the corresponding automaton state can be uniquely determined, even though the underlying POMDP state is only partially observable. The automaton state is therefore incorporated directly as part of the input to the Q-networks, together with the observation history.

The task specification is defined using Metric Interval Temporal Logic (MITL) and follows the same structural form as in the previous case study. The agent is required to visit states `$a$' and `$b$' within predefined temporal intervals. In this scenario, the temporal bounds are set to $\tau_{1} = 10$, $\tau_{2} = 15$, $\tau_{3} = 25$, and $\tau_{4} = 30$. Accordingly, the agent must visit state `$a$' between time steps 10 and 15, and subsequently visit state `$b$' between time steps 25 and 30. These intervals define strict temporal windows within which the task must be satisfied.

The reward structure is consistent with the MITL specification. The agent receives a reward of $100$ upon successfully completing both visits within their respective time bounds. Any violation, including early visits, late visits, or revisits outside the permitted intervals, results in a reward of $0$ and an immediate transition to the sink state of the corresponding Timed-LDGBA, thereby terminating the episode.

In this case study, the movement cost is set to zero, allowing the agent to focus exclusively on satisfying the temporal constraints without incurring penalties for motion. This configuration isolates the effect of partial observability and increased state-space size on task satisfaction, enabling a direct evaluation of the framework’s scalability.

\begin{figure}[H]
\centering
\subfigure[]{\resizebox*{60mm}{!}{\includegraphics{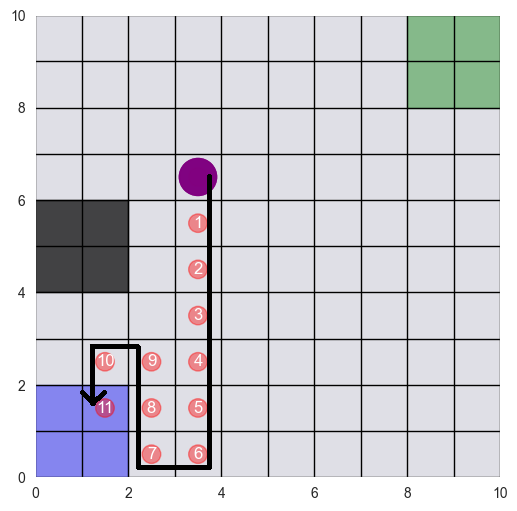}}}
\subfigure[]{\resizebox*{60mm}{!}{\includegraphics{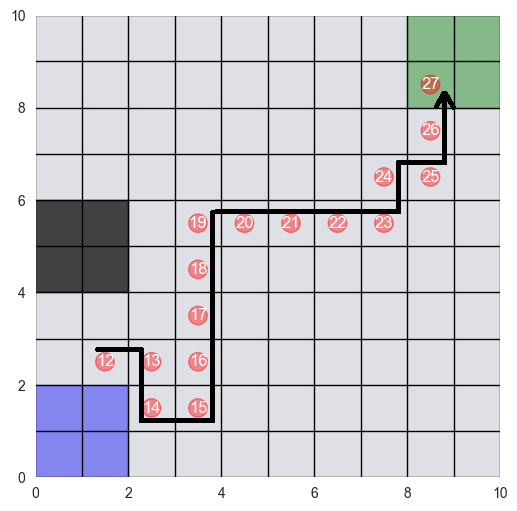}}}
\caption{ A 10 $\times$ 10 grid world environment illustrating agent trajectories:\textbf{a}) visiting label '$a$', and \textbf{b}) visiting label '$b$'. The Transition is deterministic and the reward of visiting labels is 100 with no movement penalty.}
\label{fig:10x10m0}
\end{figure}

Figure~\ref{fig:10x10m0} illustrates representative trajectories for visiting labels `$a$' and `$b$' within the specified time windows. Despite the increased environment size and partial observability, the agent successfully synthesizes policies that satisfy the MITL constraints. These results demonstrate that the proposed framework retains its ability to enforce time-bounded behaviors under partial observability and scales effectively as the planning horizon and state space increase. Although larger environments naturally introduce a higher risk of temporal violations, the agent consistently learns belief-based policies that remain within the prescribed temporal bounds.

Due to stochastic action uncertainty and observation probability, executions generated by the same policy may result in different state trajectories across runs. Consequently, even under an identical policy, the agent can exhibit variations in timing and path selection, which may occasionally lead to violations of the temporal constraints. As a result, the empirical success rate of approximately 90\% when evaluated over a large number of trials. Nevertheless, the learned policy consistently achieves a high rate of specification satisfaction, demonstrating robustness to uncertainty in both motion and sensing. Although these sources of uncertainty introduce unavoidable variability in execution outcomes, the proposed framework enables the agent to reason over temporal requirements at the policy level and to maintain reliable performance in expectation.

\subsection{Case 3: Office Scenario}

\begin{figure}
\centering
\includegraphics[width=60mm]{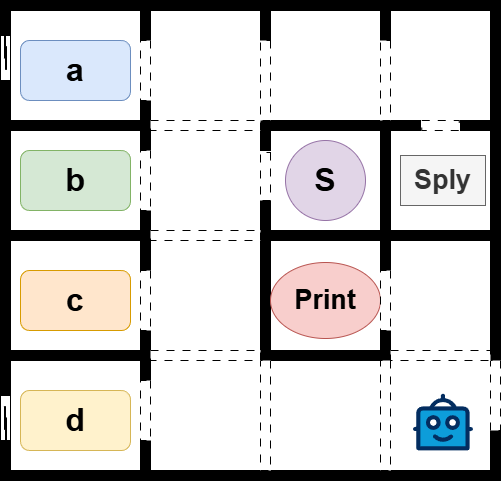}
\centering\caption{The office environment}
\label{fig:office environment}
\end{figure}

The third case concerns the office scenario. Figure \ref{fig:office environment} illustrates the virtual office environment. The workspace consists of four offices—$a$, $b$, $c$, and $d$—along with a storage room ($S$), a printer room ($Print$), and a supply station ($Sply$) used for recharging the robot (agent). In addition, Offices $a$ and $d$ each contain a large window, and the overall floor plan includes multiple doors connecting the rooms. To construct the planning model, the environment is discretized into a $4 \times 4$ grid world and modeled as a POMDP.

To account for motion uncertainty, we assume that the agent successfully executes its navigation command with probability 0.9, moving in the intended direction. With the remaining probability of 0.1, it transitions to one of the other feasible directions, selected uniformly. If an action would cause the agent to move into a wall, it remains in its current location.

In this task, the agent must first visit the printer room to collect documents and then deliver them to Office $a$. The agent is assumed to have full knowledge of its task objectives, while state information remains partially observable. Temporal constraints are specified using MITL, with bounds $\tau_{1} = 5$ and $\tau_{2} = 10$. Accordingly, the agent must visit the printer room at or after step 5, and subsequently Office $a$ at or after step 10. This task is formally expressed by the following MITL specification:
\begin{equation}
    \square\diamondsuit_{I_{1}}Print \land \square\diamondsuit_{I_{2}}a = \square\diamondsuit_{[\tau_{\text{1}}, \infty]}Print \land \square\diamondsuit_{[\tau_{\text{2}}, \infty]}a
\end{equation} 

At each state, the agent collects local observations from its surroundings in all four cardinal directions, following a fixed order: North, West, South, and East. Each observation element belongs to the set ${\text{wall}, \text{hallway}, \text{door}, \text{window}}$. It shall be noted that there is only one observation at each state, but multiple states may have identical observation patterns. For example, Offices ‘$b$’ and ‘$c$’ share the same observation, i.e., $o(b) = o(c) = {\text{wall}, \text{wall}, \text{wall}, \text{door}}$. As a result, the observation space $O$ of this POMDP consists of 13 distinct observations.

The reward structure is designed to strictly enforce the MITL constraints. If the agent visits the required labeled states within the designated temporal bounds, it receives a reward of $100$. Any violation, including early arrivals, late arrivals, or invalid revisits, yields a reward of $0$ and results in an immediate transition to the sink state of the corresponding Timed-LDGBA, thereby terminating the episode.

\begin{figure}[H]
\centering
\subfigure[]{\resizebox*{60mm}{!}{\includegraphics{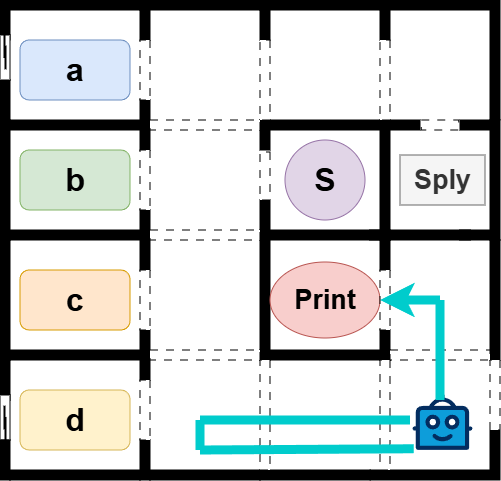}}}
\subfigure[]{\resizebox*{60mm}{!}{\includegraphics{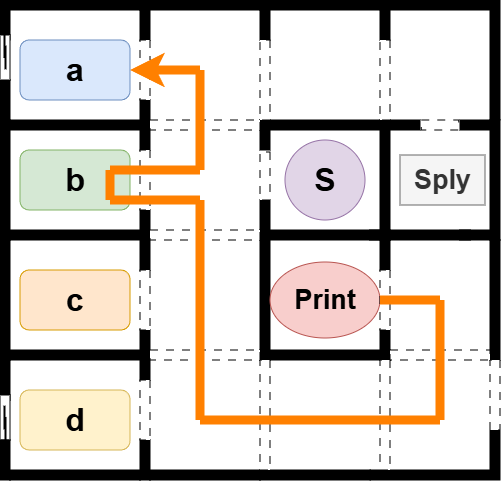}}}
\caption{The office scenario: \textbf{a}) visiting printer room, and \textbf{b}) visiting office `$a$'.}
\label{fig:office}
\end{figure}

After the training process converges, the optimal policy is derived from the Q-networks. The resulting trajectories are plotted in Figure \ref{fig:office}, shown as straight-line paths for clarity in illustrating successful task completion. It can be observed that, upon leaving the initial state, the agent moves toward the printer room, depicted by the blue path. After arriving at the printer room and collecting the documents, the agent leaves and proceeds to Office $a$ along the orange path to complete the delivery.

By embedding timed automaton dynamics into a POMDP model, the approach enables learning under uncertainty regarding both system dynamics and state observability. The successful navigation from the initial location to the printer room, followed by the timed delivery to Office $a$, demonstrates that the agent can learn policies that integrate belief updates, stochastic transitions, and strict temporal logic requirements. 

\section{Discussion}

This study investigates the integration of MITL with RL in order to address robotic tasks that demand both correct sequential behavior and strict temporal guarantees.  The study contributes to this growing research direction by constructing a unified framework in which MITL specifications are translated into a Timed Timed-LDGBA. By synchronizing the resulting automaton with either a MDP or a POMDP, we construct product timed models that are directly amenable to Q-learning. This formulation enables agents to reason over multiple time-bounded objectives simultaneously, while a simple yet expressive reward structure penalizes temporal violations and guides learning toward satisfaction of all MITL constraints.

The effectiveness of the proposed framework is demonstrated across three simulation environments: a $5 \times 5$ grid-world modeled as an MDP, a larger $10 \times 10$ grid-world modeled as a POMDP, and an office-like service-robot scenario. Collectively, the results highlight several key insights. First, even under stochastic motion dynamics, agents consistently learn policies that satisfy the specified timing windows, indicating that the product timed MDP/POMDP formulation successfully exposes temporal structure to the learning process. 

In the $5 \times 5$ MDP environment, the agent reliably satisfies recursive, time-bounded visitation tasks involving labeled states ($a$ and $b$). When movement penalties are introduced, the agent adapts its behavior by shifting from exploratory trajectories to more efficient paths, while still preserving all temporal requirements. Scalability is examined in the $10 \times 10$ POMDP environment, where the state space and planning horizon are significantly larger. Despite these challenges and the presence of partial observability, the agent continues to satisfy the MITL constraints, illustrating that the approach remains effective as problem complexity increases. The office scenario further highlights the strengths of the proposed method in complex environments. By embedding timed-automaton dynamics into a POMDP, the agent is able to integrate belief updates, stochastic transitions, and strict temporal logic requirements. 

While the presented results are encouraging, several limitations and opportunities for refinement remain. In the current formulation, the T-LDGBA relies on a single global clock to track the progression of time across all temporal objectives. As a result, all time intervals in the MITL specification are evaluated with respect to the same global notion of time, which may limit expressiveness when tasks require independent or overlapping temporal windows associated with different events. Moreover, as the number of time-bounded objectives increases, the size of the automaton and the corresponding product MDP or POMDP can grow rapidly, leading to increased computational and memory demands. Future work may therefore explore extensions that support multiple local time references, as well as abstraction, decomposition, or hierarchical learning strategies to improve scalability for large and complex tasks.

Finally, the proposed framework naturally lends itself to several promising extensions. One important direction is the investigation of multi-agent settings, where multiple agents must satisfy individual or coupled time-interval-based specifications while operating in shared and potentially partially observable environments. Many real-world applications—such as cooperative robotics, intelligent transportation systems, and human–robot collaboration—require coordinated satisfaction of timed objectives across agents. Integrating MITL-based specifications with multi-agent reinforcement learning, while preserving temporal guarantees under uncertainty, would represent a meaningful and impactful extension of this work.

\section*{Declaration of Generative AI and AI-assisted technologies in the writing process}
During the preparation of this work, the authors used ChatGPT in order to improve the readability and language during the writing process. After using this tool/service, the authors reviewed and edited the content as needed and took full responsibility for the content of the publication. 

\section*{Declaration of Competing Interest}
The authors declare that they have no known competing financial interests or personal relationships that could have appeared to influence the work reported in this paper.

\bibliographystyle{ieeetr}
\bibliography{bib}
\end{document}